# Swin Transformer coupling CNNs Makes Strong Contextual Encoders for VHR Image Road Extraction

*Tao Chen, Yiran Liu, Haoyu Jiang, Ruirui Li\**

Beijing University of Chemical Technology, College of Information Science & Technology
ilydouble@gmail.com

## ABSTRACT

Accurately segmenting roads is challenging due to substantial intra-class variations, indistinct inter-class distinctions, and occlusions caused by shadows, trees, and buildings. To address these challenges, attention to important texture details and perception of global geometric contextual information are essential. Recent research has shown that CNN-Transformer hybrid structures outperform using CNN or Transformer alone. While CNN excels at extracting local detail features, the Transformer naturally perceives global contextual information. In this paper, we propose a dual-branch network block named ConSwin that combines ResNet and SwinTransformers for road extraction tasks. This ConSwin block harnesses the strengths of both approaches to better extract detailed and global features. Based on ConSwin, we construct an hourglass-shaped road extraction network and introduce two novel connection structures to better transmit texture and structural detail information to the decoder. Our proposed method outperforms state-of-the-art methods on both the Massachusetts and CHN6-CUG datasets in terms of overall accuracy, IOU, and F1 indicators. Additional experiments validate the effectiveness of our proposed module, while visualization results demonstrate its ability to obtain better road representations.

## 1. INTRODUCTION

Extracting roads from Very High Resolution (VHR) Remote Sensing Imagery (RSI) is vital for applications such as updating Geographic Information Systems (GIS), urban planning, vehicle navigation, and disaster assessment. Despite decades of research, no satisfactory automated solution to this task has been found due to several challenges that limit extraction accuracy, including shadows, occlusions caused by trees, buildings, etc., and intra-class variance of road surfaces. With the rapid growth of available data and computing power, deep learning techniques have achieved breakthroughs in various computer vision tasks. Given that road extraction can be regarded as a binary semantic segmentation problem, researchers have employed different Convolutional Neural Network (CNN) architectures for road segmentation.

U-Net [2] is a popular network architecture used for road segmentation, which leverages symmetrical encoding and decoding structures to restore pixel-level road information. Dilated convolution is another effective technique used to increase the receptive field without decreasing the resolution of feature maps. D-LinkNet [3], which won first place in the DeepGlobe 2018 Road Extraction Challenge, is built on the LinkNet architecture with a dilated convolutional layer at its center. Other studies [4] have attempted to construct global context using dilated convolutional neural networks. Despite significant progress made by these methods, there is still room for improvement in terms of the completeness and connectivity of results, as shown in Figure 1(d).

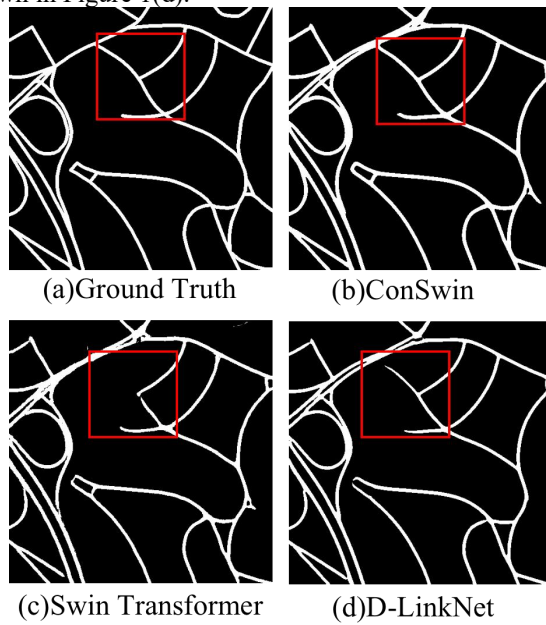

(a)Ground Truth    (b)ConSwin
(c)Swin Transformer    (d)D-LinkNet

Figure.1 An example of road segmentation task

Roads in remote sensing images possess unique characteristics, including elongated coherent regions[1] with similar spectra and textures, a fixed width, limited curvature, and other specific features. Accurately perceiving these characteristics and modeling geometric context is crucial for improving the integrity and connectivity of road segmentation. Attention mechanisms enable deep neural networks to selectively focus on relevant information while ignoring irrelevant information. The Transformer model is exclusively based on the attention mechanism and does not use recurrence or convolutions. Recently, the Vision Transformer (ViT) and its variant Swin Transformer [5] have demonstrated multiple advantages for various vision tasks, including design simplicity, robustness, and state-of-the-art performance. As Transformers measure the relationship between input token pairs through attention, they inherently possess the ability to perceive global context.

Vision Transformers lack the inherent inductive bias of convolutional neural networks, which can be effectively trained on smaller datasets. The success of Vision Transformers mainly relies on pre-training on large-scale datasets, making them less directly adaptable to small-scale datasets such as some RSI datasets for road extraction. According to [6], models with self-attention or convolutions each have their limitations. While Transformers are skilled in modeling long-range global contexts, they may not be proficient in extracting fine-grained local feature patterns (as shown in Figure 1(c)). Conversely, Convolutional Neural Networks (CNNs) excel at exploiting local information and have become the de facto computational block for many computer vision tasks.

To address the lack of inductive bias and insufficient awareness of local features, this paper proposes a convolution-augmented Swin Transformer structure for learning. It combines the advantages of Swin Transformers and CNNs to make a stronger context encoder . It tries to fetch long-range dependencies and local features interactively during the multiple encoding stages, thereby improving the integrity and connectivity of road segmentation (see Figure1(b)). Below, we summarize the key contributions of this work:

1.We introduce a novel network block called ConSwin, which integrates Swin Transformers with CNNs by utilizing the hyperbolic tangent function to align features outputted by Swin Transformer with CNN.

2.We construct an encoder-decoder network based on ConSwin blocks and propose two enhanced connection structures. One enhances geometric orientation features, while the other enhances texture features. These two connection structures enable the network to more effectively capture both local and global contextual features of roads.

3.We perform extensive experiments on two datasets: Massachusetts and CHN6-CUG. The results show that our proposed method can achieve more complete and accurate road segmentation outcomes. Further model analyses demonstrate the effectiveness of our proposed method compared to other road segmentation methods.

## 2. RELATED WORK

### 2.1 Deep convolution-based VHR Road extraction

Early remote sensing road extraction is a collection of image classification tasks to identify whether the images contain a road, which can be competent by convolutional networks. With the development of semantic segmentation, road extraction is also refined to the pixel level, and variants of U-Net become the primary choice. D-LinkNet inherits the skip connection and encoder-decoder structure from U-Net, while using dilated convolution instead of pooling layers for downsampling. Then more spatial information can be retained. Recent studies have shown that contextual information is essential to improve the accuracy of semantic segmentation, Deeplab [24] and Dilated Conv [25] proposed the atrous convolution to expand the network receptive field without sacrificing resolution, which allows the network to acquire contextual information over a larger field. Based on this, an atrous spatial pooling pyramid (ASPP) module was developed to include contextual information at multiple scales, thus increasing the receptive field of the network.

### 2.2 Vision Transformer for VHR RSI

Transformer [18], a class of networks based on the self-attention mechanism, was first used for machine translation and subsequently performed well in many NLP tasks. To make the Transformer applicable to computer vision tasks, ViTs [19] segment images into small patches which correspond to "tokens" in NLP, and extend the positional coding to preserve the positional relationships between patches. ViTs show impressive performance in a series of computer vision tasks. Compared to traditional CNN networks, the self-attentive mechanism in ViTs can capture and aggregate global features of images more effectively. Recent studies in [20][21] have shown that ViTs have the ability to model global contextual interactions and the flexibility to adjust the modeling ability of different regions to counteract noise in the data and learn effective feature representations. In [22], a Pyramid Visual Transformer (PVT) architecture was used to extract high-resolution multi-scale feature maps for performing pixel-level intensive prediction tasks. Compared to convolutional backbones with a comparable number of parameters, PVT can be more capable of extracting globalized information about the context, which significantly improves the performance of the model on the road segmentation task.

### 2.3. The combination of Transformer and CNN

Although the global information aggregation capability of the Transformer framework has led to significant growth in many tasks on remotely sensed images, however, convolutional neural networks still have advantages for feature extraction of local detail information. Recent research on road extraction has developed a combination of Transformer and CNN. These methods have both the capability of Transformer for contextual information extraction and CNN for local detail information extraction. Efficient-Transformer [23] introduces a lightweight Transformer framework incorporating an implicit edge enhancement technique. TransUNet uses a hybrid CNN-Transformer model, which first maps the original image into a feature map using CNN and then feeds the patches from the feature map into Vision Transformer for a second encoding. The Transformer's ability to encode global contexts is employed to achieve a better representation. BDTNet and Roadformer are two recent works on road extraction. They extended the Swin transformer and add decoder structures to better reconstruct roads. In order to maximize the advantages of CNN to extract local features and Transformer to extract global context, the authors of Conformer proposed a dual-path structure so that the information in their learning process can be interactive. Experiments show that the CNN-Transformer hybrid encoder performs better than using only Transformers as an encoder.

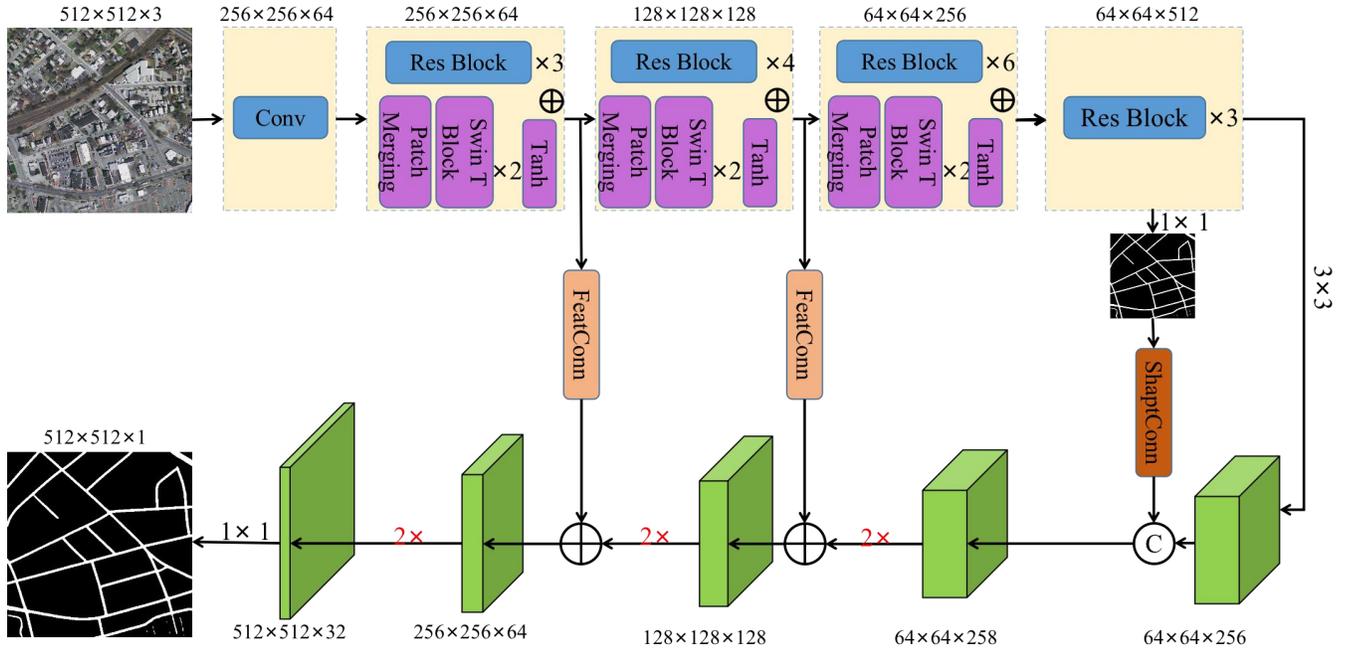

Figure 2 The network for VHR RSI road extraction.

## 3. PROPOSED METHOD

### 3.1 Road extraction network

The road segmentation task demands an understanding of both local details and global context. Prior to the emergence of Vision Transformers (ViTs), the hourglass network was one of the most successful frameworks for road segmentation in remote sensing images. However, as ViTs gained popularity in computer vision tasks, researchers discovered that using ViTs alone to encode remote sensing images was less effective than combining CNNs with ViTs. As a result, some recent studies have attempted to incorporate ViT structural blocks into the hourglass network or organize them into a pyramid structure similar to better extract roads with higher accuracy.

This paper also follows the idea of combining ViTs and CNNs to design a road extraction network. We propose a new encoding module called ConSwin that integrates the advantages of both Swin Transformer and CNN. Unlike TransUNet, which inputs CNN features as tokens into the transformer or RoadFormer, which builds encoders only with transformers, our approach employs ConSwin modules to construct the encoding part of the hourglass network. The overall network architecture is shown in Figure 2, featuring an hourglass-shaped symmetrical structure with a downsampling encoding path, an upsampling decoding path, and intermediate connections. The downsampling encoding path comprises three connected ConSwin modules that are critical components of our method for obtaining local features and global contexts. The decoders in the upsampling decoding path are designed similarly to U-Net. In the middle of the network, we adopt a shape-augmented connection structure called shapConn to collect road direction information and apply it to the feature map to guide road extraction. Additionally, we use a feature-enhanced connection structure called FeatConn between the encoder and decoder at the same level to transfer local detail statistics to the decoder for better reconstruction of segmentation results. In the following subsections, we will provide a detailed explanation of the ConSwin module (3.2) and introduce the two connection structures (3.3 and 3.4), respectively.

### 3.2. ConSwin Module

A recent comparative study between Vision Transformer models (ViT) and ResNet revealed that ViT is more prone to misidentifying objects with similar shapes, whereas ResNet is more likely to rely on texture information. Google researchers [6] conducted an additional analysis of the internal representation structure of ViT and ResNet, revealing significant differences between the two architectures. Firstly, ViT incorporates more global information than ResNet at lower layers. Secondly, incorporating local information at lower layers remains crucial for achieving high accuracy.

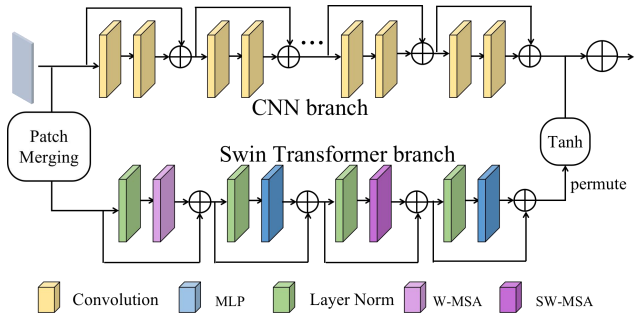

Figure.3 The structure of ConSwin Module

Building upon the aforementioned findings, this paper introduces a novel feature encoding module, ConSwin (Figure.3), that leverages both Transformer and ResNet concepts. The module features a dual-branch structure consisting of ResNet blocks in one branch and two successive Swin Transformer blocks in the other.

Swin Transformer is a Vision Transformer variant that utilizes a hierarchical design with downsampling and a shifted window mechanism. It replaces the standard multi-head self-attention module (MSA) in the Transformer block with the window multi-head self-attention module (W-MSA). Each Swin Transformer block also contains two Layer Norm (LN) layers and a Multi-Layer Perception (MLP) with Gaussian Error Linear Unit (GELU) nonlinearity. Figure 3 illustrates that when we use two consecutive Swin Transformer blocks, the W-MSA module in the latter block is replaced by the shifted window multi-head self-attention module (SW-MSA).

We use the PatchMerging operation to transform the image of feature map into patch tokens, which is presented in Figure 4. It first selects elements at intervals of 2 along rows and columns, akin to convolution/pooling with a stride of 2. These selected elements are then concatenated to create a tensor. As interval sampling reduces the length and width of the original feature map by half, the channel dimension becomes four times the original dimension. Subsequently, the Patch Merging module adjusts the number of channels through downsampling to establish a layered design.

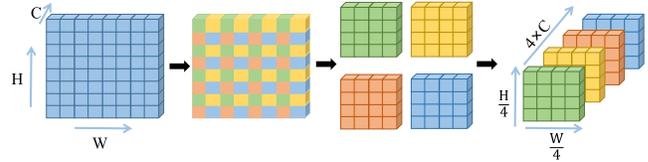

Figure 4 The processing of patch merging.

During the learning process, there is a notable inconsistency between CNN and Transformer with regards to feature interest and magnitude, making it difficult to effectively fuse the two methods. To quantify the difference in feature magnitude, a dimensional alignment of two branch features via the permutation operation is performed, followed by an averaging of the absolute value of the difference on the feature maps.

The statistical differences in magnitude of features is calculated by:

$$S_i = \frac{1}{N}\sum_{j=1}^{N} abs(x_j - y_j) \quad (1)$$

where $S_{i}$ is the statistical difference of the i-th sample, N is the number of features, $x_{j}$ is the j-th feature of ResNet, and $y_{j}$ is the j-th feature of the Swin Transformer.

In Figure 5, the blue curve shows the statistical results for 100 instances randomly chosen, indicating a relatively large statistical difference between the features obtained by the ResNet branch and the Swin transformer, with fluctuating curves of the differences. Directly adding these two sets of features for feature fusion results in the Transformer branch quickly dominating the training process, thereby hindering the functionality of ResNet and the model convergence. This phenomenon is mainly due to the fact that Swin Transformers uses Layer Norm normalization and obtains larger eigenvalues. In comparison, the eigenvalues of ResNet are always less than 1.

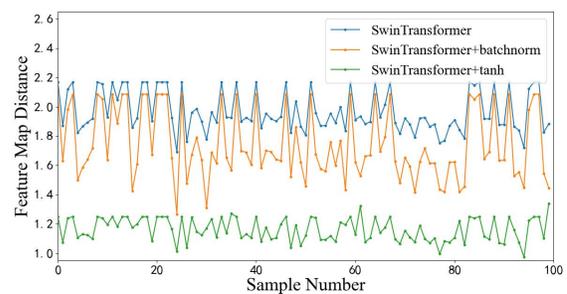

Figure.5 the statistical results for 100 instances randomly chosen which are counted by Equation (1)

Even using layer Norm normalization again cannot reduce the magnitude difference between the two branch features. The orange curve in Figure 5 shows the statistics after the Layer Norm operation for the Swin transformer. The results demonstrated that the using of Layer Norm did not significantly reduce the gap.

To achieve better feature fusion between the two branches, we apply the hyperbolic tangent transformation to the Swin Transformer features. This technique effectively compresses the scale of difference. The green line in Figure 5 illustrates the statistics between features of ResNet and those output by Swin Tranformer and transformed by the hyperbolic tangent function, with a minimal gap facilitating superior feature fusion. Let z_{j} be the j-th feature of output map, it can be calculated by :

$$z_j = x_j + tanh(y_j) \quad (2)$$

where tanh() is the hyperbolic tangent function.

### 3.3. Feature-enhanced connection

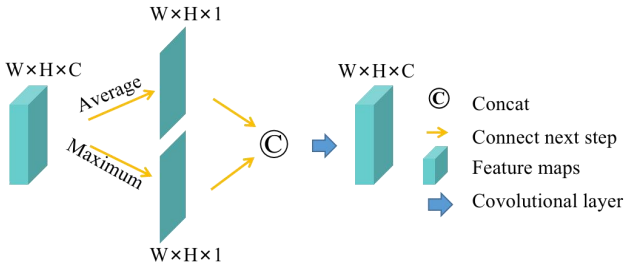

Figure.6 FeatConn structure

The proposed network structure effectively learns both local details and global contexts relevant to road extraction. Nevertheless, to achieve better reconstruction of road segmentation results, more detailed information is necessary. Connections between encoders and decoders transfer shallower features preceding down-sampling to the corresponding decoders, aiding in recovering high-resolution feature maps.

Prior work integrated attention structures into connections within the hourglass framework to filter out irrelevant features, whereas Transformer-based encoding structures such as PVT directly connect shallow encoding features to the decoder. However, neither design is well-suited for the network framework proposed in this paper. To transmit detailed information more efficiently, we introduce a statistical information-based feature-enhanced connection structure. Figure 6 depicts the structure, which separates the average and maximum values of the feature maps and concatenates them together.

Assuming the input is X, the output can be expressed as:

$$Y = Conv(Concat(Avg(X), Max(X)))$$

Avg() and Max() represent the average value and the maximum value along channels respectively, and Concat() is the concatenation operation.

### 3.4. Shape-augmented connection

Previous studies have demonstrated that the orientation information of roads can aid in guiding road segmentation and improve connectivity results [26]. Thus, this paper introduces a connection structure called shapConn in the middle of the hourglass network to extract and fuse road direction features. First, we generate a low-resolution road segmentation map by applying a BCE loss function to a 1x1 convolution. We then compute the Sobel components on the map along the x and y directions as directional features. Figure 7 illustrates this connection structure.

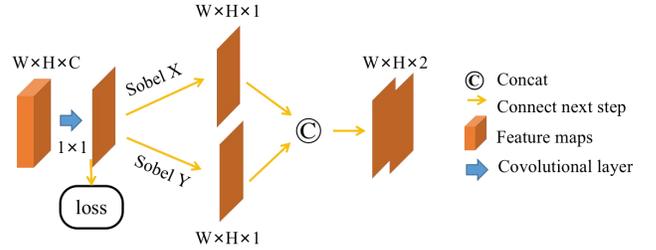

Figure.7 ShapConn structure

Assuming that X represents the input and X' denotes the segmentation map, the output can be expressed as follows:

$$Y = Concat(S_x * X', S_y * X')$$

with

$$S_y = \begin{bmatrix} -1 & -2 & -1 \\ 0 & 0 & 0 \\ 1 & 2 & 1 \end{bmatrix}, S_y = \begin{bmatrix} -1 & 0 & 1 \\ -2 & 0 & 2 \\ -1 & 0 & 1 \end{bmatrix}$$

and Concat() is the concatenation operation and * depicts the convolution operation.

### 3.5. The loss function

We establish a road semantic segmentation network for extracting roads, where the road coverage is typically much smaller than that of the background. This can result in an imbalanced dataset and under-extraction of roads. To mitigate this issue, we train the road segmentation model utilizing weighted cross-entropy loss, which is expressed as follows:

$$L_1 = \frac{1}{N}\sum_{i=1}^{N}((1 + \alpha \cdot w_i)y_i \cdot \log(y_i') + (1 - y_i)\log(1 - y_i')) + L_{smooth} \quad (8)$$

where $y_i$ represent predicted value, $y_i'$ represent the Ground Truth, N represents number of total pixels, $w_i$ is set by $1 - \frac{number\ of\ road\ pixels}{number\ of\ total\ pixels}$, and $\alpha$ is a constant hyperparameter. We also add the smooth-L1 regularization term to the loss function to stabilize the training of the network.

Within the network's intermediate connection structure, we have incorporated a BCE loss to acquire a preliminary road segmentation outcome. Thus, the overall loss function primarily comprises two components: $L_1$, representing the loss of the final segmentation result, and $L_2$, reflecting the loss of the intermediate segmentation result. These two training objectives entail varying degrees of refinement, necessitating dynamic weighting to ensure optimal learning of parameter models by the network. Nonetheless, manual weight adjustment during the training process can be challenging. Drawing inspiration from prior work [29], we employ the maximized Gaussian likelihood approximation to balance the two losses, as follows:

$$L_T \approx \frac{1}{\sigma_1^2}L_1 + \frac{1}{\sigma_2^2}L_2 + \ln\sigma_1\sigma_2 \quad (11)$$

Among them, $\sigma_1$ and $\sigma_2$ represent the standard deviation. The larger the standard deviation, the smaller the weight of this loss. The magnitude of these two standard deviations is estimated by an exponential transformation of the two learnable parameters.

## 4. EXPERIMENTS AND ANALYSIS

### 4.1. Experimental setup

The experiments were conducted on an Ubuntu 16.04 platform with an NVIDIA 1080Ti. To further improve the generalization, we used data augmentation techniques such as random horizontal rotation, random vertical rotation, random scaling, and random hue-saturation-value during the model training process. The Adam optimizer was used with an initial learning rate of 0.0002, and the network was trained using a batch size of 2 or 3. The binary output was generated by averaging each prediction over the augmented image.

Two datasets, Massachusetts road dataset [10] and CHN6-CUG road dataset [17], were selected. They were captured from different regions of roads, each containing urban and rural roads with complex and diverse data. The spatial resolution of the Massachusetts road dataset is 1.2m, while the CHN6-CUG road dataset can reach a resolution of 50cm per pixel. The significant difference in resolution allows for a better evaluation of the generalization of the model.

The Massachusetts road dataset was created by Mihn and is an aerial imagery dataset that includes 1108 images for training, 14 images for validation, and 49 images for testing. Each image contains 1500 × 1500 pixels. They were resized to 512 x 512 pixels in our experiments.

The CHN6-CUG road dataset was generously provided by China University of Geosciences in Wuhan, China. This dataset consists of a large-scale collection of satellite images from representative cities in China, including the Chaoyang area of Beijing, the Yangpu District of Shanghai, Wuhan city center, the Nanshan area of Shenzhen, the Shatin area of Hong Kong, and Macao, each with varying levels of urbanization. Each image in the dataset has a size of 1024 × 1024 pixels, which is then scaled down to 512 × 512 pixels to maintain consistency with the DeepGlobe dataset. In total, there are 4511 labeled images of size 512 × 512, which are divided into 3608 images for training and 903 for testing.

### 4.3 Evaluation Metrics

In order to perform a comprehensive evaluation of the considered methods, in this experiment, we use Precision, Recall, F1-Value (F1), Intersection over Union (IOU), and Overall Accuracy (OA). These are the most widely used metrics in binary segmentation tasks[30]. They are all calculated based on TP, FP, TN, and FN. TP denotes correctly identified positive samples, FP denotes incorrectly identified positive samples, TN denotes correctly identified negative samples, and FN denotes incorrectly identified negative samples. These evaluation metrics are defined by:

- The precision metric measures the accuracy of positive predictions in a given sample. It is calculated using the following formula:

$$\text{Precision} = \frac{TP}{TP + FP} \quad (12)$$

- The recall metric measures the proportion of true positive samples that are correctly identified by the model. It is defined by the following formula:

$$\text{Recall} = \frac{TP}{TP + FN} \quad (13)$$

- F1 is denoted F1-Score, it is a combined evaluation metric that takes into account both missed detections and false detections for samples with positive true labels. Its definition formula is as follows:

$$F1 = 2 \times \frac{\text{Precision} \times \text{Recall}}{\text{Precision} + \text{Recall}} \quad (14)$$

- IOU stands for Intersection over Union, it is a metric used to evaluate the overlap between the predicted and ground truth degree of overlap. The formula for IOU is as follows:

$$\text{IOU} = \frac{TP}{TP + FP + FN} \quad (15)$$

- OA is the Overall Accuracy, it represents the percentage of correctly predicted samples out of all samples. The formula for calculating it is as follows:

$$OA = \frac{TP + TN}{TP + FP + FN + TN} \quad (16)$$

### 4.4 Experiments

In this section, we report experimental results that demonstrate the efficacy of our proposed method in road extraction. Specifically, we present a comparative analysis with state-of-the-art methods such as U-Net, Deeplab v3+[16], D-LinkNet, SwinT ransformer, PVT, TransUNET[9], and Roadformer, which are based on CNNs, ViTs, and hybrid CNN-Transformer architectures.

TABLE I
COMPARISON ON THE MASSACHUSETTS ROAD DATASET

| model | backbone | Precision(%) | ReCall(%) | F1(%) | IOU(%) | OA(%) |
|---|---|---|---|---|---|---|
| UNet | \ | 79.18 | 71.61 | 75.21 | 60.27 | 97.78 |
| Deeplabv3+ | ResNet34 | **83.33** | 71.37 | 76.88 | 62.45 | 97.98 |
| D-LinkNet | ResNet34 | 82.53 | 71.67 | 76.71 | 62.22 | 97.95 |
| Swin Transformer | Swin-T | 83.26 | 73.33 | 77.99 | 63.92 | 98.05 |
| PVT | PVT-S | 82.01 | 70.67 | 75.92 | 61.19 | 97.89 |
| TransUNet | \ | 82.45 | 71.22 | 76.43 | 61.85 | 97.93 |
| RoadFromer | Swin-T | 79.43 | 76.79 | 78.09 | 64.04 | 97.97 |
| ours | ResNet34 | 81.11 | **79.17** | **80.13** | **66.84** | **98.15** |

TABLE II
COMPARISON ON THE CHN6-CUG ROAD DATASET

| model | backbone | Precision(%) | ReCall(%) | F1(%) | IOU(%) | OA(%) |
|---|---|---|---|---|---|---|
| UNet | \ | **81.39** | 64.85 | 72.18 | 56.48 | 97.13 |
| Deeplabv3+ | ResNet34 | 81.05 | 66.31 | 72.94 | 57.41 | 97.18 |
| D-LinkNet | ResNet34 | 80.98 | 62.00 | 70.23 | 54.12 | 96.99 |
| Swin Transformer | Swin-T | 79.98 | 66.15 | 72.41 | 56.76 | 97.11 |
| PVT | PVT-S | 72.63 | 38.78 | 50.56 | 33.83 | 95.65 |
| TransUNet | \ | 78.84 | 60.85 | 68.69 | 52.31 | 96.82 |
| RoadFromer | Swin-T | 67.39 | 58.16 | 62.43 | 45.38 | 95.99 |
| ours | ResNet34 | 78.91 | **75.60** | **77.22** | **62.89** | **97.44** |

Experimental results on the test sets are presented in Tables 1 and 2, with bold values indicating the best performance across various evaluation metrics. The superiority of our proposed model over other methods is evident from its higher scores for F1, IOU, and OA.

On the Massachusetts road dataset, our model outperforms the second-best Swin Transformer by 2.18% on F1, 2.92% on IOU, and 0.1% on OA, achieving respective scores of 80.13%, 66.84%, and 98.15%. Similarly, on the CHN6-CUG road dataset, our model surpasses the second-best Deeplab v3+ by 5.28% on F1, 5.47% on IOU, and 0.26% on OA, with respective scores of 77.22%, 62.89%, and 97.44%. Above results demonstrate that the model proposed in this study exhibits the highest overall accuracy and comprehensive ability for detecting road information in remote sensing images, with minimal missed detections of road pixels. While the Swin Transformer achieved good accuracy on the Massachusetts dataset, it did not perform as well as Deeplab v3+ on the higher resolution CHN6-CUG remote sensing dataset. This observation may be attributed to the fact that road images contain intricate details, which can be challenging for transformer-based networks to perceive, given their relative weakness in processing detailed information. To further evaluate the effectiveness of our experimental design in handling road occlusions and complex road patterns, we selected road scenes with tree occlusions, building shadows occlusions, and complex and blurry road patterns for analysis. The results are presented in Figures 8 and 9.

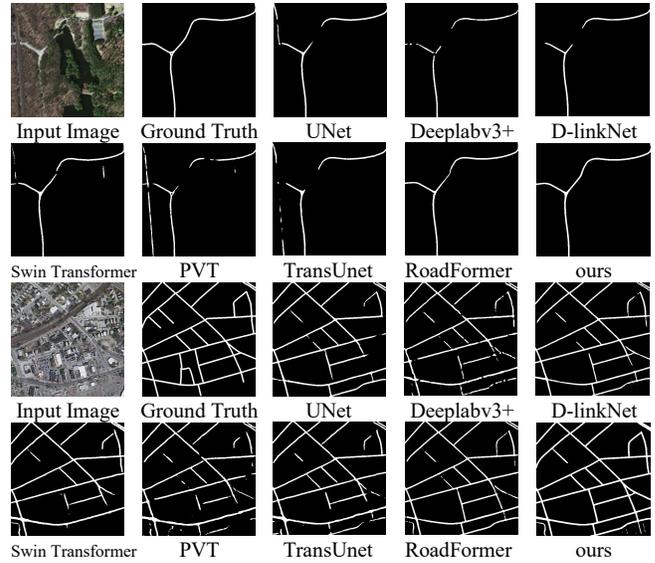

Figure .8 Sample results of different models on Massachusetts.

As demonstrated by the first example in Figure 8, our method can accurately extract roads even in areas with dense tree cover. In comparison to the Swin Transformer, the road extracted by our proposed method is continuous without any interruptions or mis-extractions.

In the second example of Figures 8, the roads intersect, and the border of the road image in the lower right corner is indistinct. D-LinkNet can segment roads with blurred boundaries but produces disconnected results. RoadFormer can also segment roads with blurred boundaries while preserving continuity, but it is unable to extract interrupted roads. Our proposed method achieves road segmentation results that are most similar to the Ground Truth when compared to the aforementioned methods.

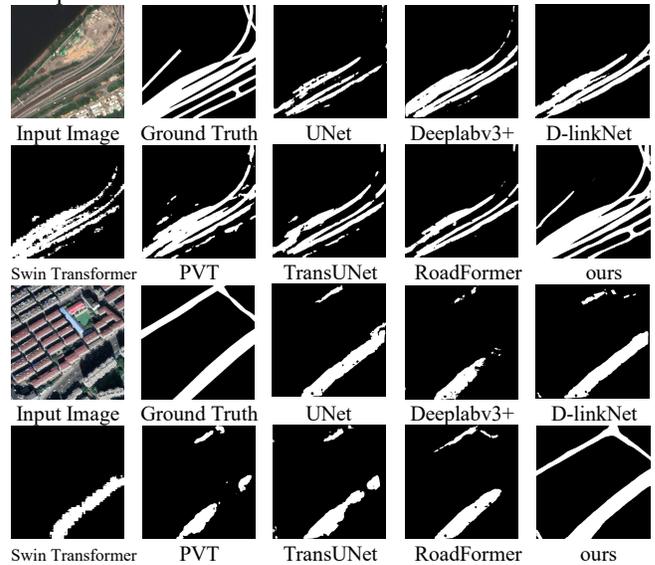

Figure.9 Sample results of different models on CHN6-CUG.

Figure 9 displays road extraction results on the CHN6-CUG dataset, featuring two samples. In the first sample, roads with diverse shapes are arranged in a row, and precise road segmentation requires accurate perception of details such as texture and local shape. Our proposed method significantly outperforms other methods in terms of segmentation accuracy for this sample. In the second sample, multiple connected roads are occluded by shadows, necessitating attention to both local details and overall structural relationships among the roads to achieve accurate segmentation. Once again, our proposed method exhibits significant improvement over other methods in this scenario. Our proposed method is capable of attending to both local details and global semantics. By utilizing an hourglass-shaped network and an improved geometry and texture connection structure, our approach can capture information related to road extraction to the fullest extent possible, thereby enabling more accurate extraction of road information in complex scenes.

To gain a better understanding and demonstrate the effectiveness of each module in our experiments, we conducted ablation experiments using the test dataset. As shown in Tables 7 and 8, the road extraction performance of each model on both the Massachusetts and CHN6-CUG datasets improved. The ConSwin module effectively extracts road semantic information, while the two connection structures, FeatConn and shapConn, better preserve road information to restore road segmentation results during subsequent decoding processes. Overall, combining all modules together leads to the best performance in terms of road segmentation accuracy.

TABLE III
COMPARISON ON THE MASSACHUSETTS ROAD DATASET

| Model | | | Massachusetts road dataset | | |
|---|---|---|---|---|---|
| ConSwin | ShapConn | FeatConn | F1(%) | IOU(%) | OA(%) |
| | | | 78.37 | 64.43 | 98.00 |
| √ | | | 79.20 | 65.56 | 98.08 |
| | √ | | 79.53 | 66.21 | 98.08 |
| | | √ | 78.44 | 64.53 | 98.00 |
| √ | √ | | 79.85 | 66.46 | 98.14 |
| √ | | √ | 79.81 | 66.40 | 98.14 |
| | √ | √ | 79.99 | 66.66 | 98.14 |
| √ | √ | √ | 80.13 | 66.84 | 98.15 |

TABLE IV
COMPARISON ON THE CHN6-CUG ROAD DATASET

| Model | | | CHN6-CUG road dataset | | |
|---|---|---|---|---|---|
| ConSwin | ShapConn | FeatConn | F1(%) | IOU(%) | OA(%) |
| | | | 69.55 | 55.31 | 96.65 |
| √ | | | 75.78 | 60.01 | 97.36 |
| | √ | | 75.64 | 60.97 | 97.37 |
| | | √ | 71.17 | 55.25 | 97.10 |
| √ | √ | | 76.17 | 61.52 | 97.36 |
| √ | | √ | 76.29 | 61.67 | 97.34 |
| | √ | √ | 76.22 | 61.64 | 97.34 |
| √ | √ | √ | 77.22 | 62.89 | 97.44 |

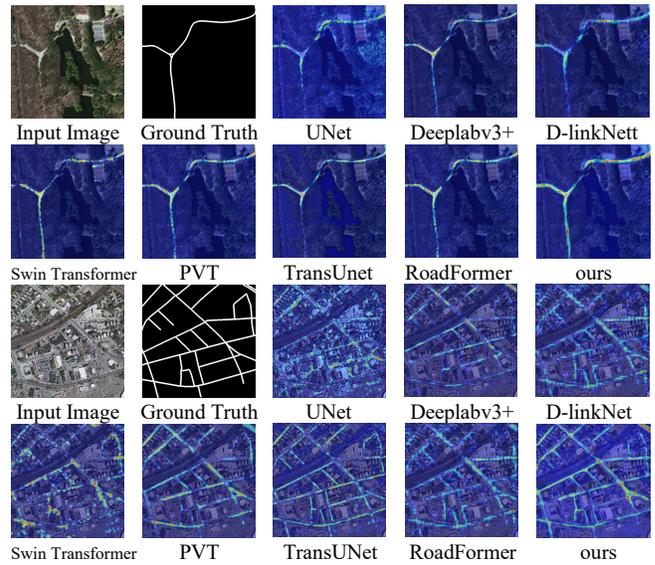
Fig.10 Visualized heat map of different models on Massachusetts.

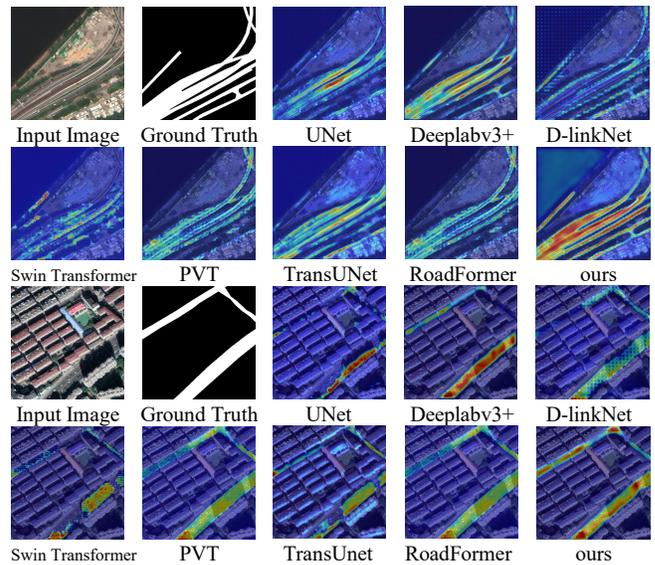
Fig.11 Visualized heat map of different models on CHN6-CUG.

To investigate the semantic information captured by each model for road information, this paper employed visualization heatmaps [31], which are presented in Figures 10 and 11. Our method effectively concentrates on road regions and related features. The figures illustrate that our model can capture information from obscured and complex regions of the road, indicating that our proposed approach can effectively handle the challenges posed by road occlusion and complexity.

## 4. CONCLUSION

This paper focuses on improving Very High-Resolution Remote Sensing Image (VHR RSI) road segmentation. To this end, we propose ConSwin, a novel feature encoding module that effectively models both local details and global context for improved representation of roads. Additionally, we introduce an hourglass-shaped network based on ConSwin, which includes two innovative connection structures designed to enhance road extraction performance. Experimental results demonstrate that our proposed method achieves the best road segmentation accuracy by improving road integrity and connectivity.